\documentclass[10pt,twocolumn,letterpaper]{article}

\usepackage{cvpr}              

\usepackage{graphicx}
\usepackage{amsmath}
\usepackage{amssymb}
\usepackage{booktabs}
\usepackage{epstopdf}
\usepackage[pagebackref=true,breaklinks=true,letterpaper=true,colorlinks,bookmarks=false]{hyperref}
\usepackage{booktabs}
\usepackage{stfloats}
\usepackage{float}
\usepackage{multirow}
\usepackage{bbding}
\usepackage{bm}
\usepackage{amsfonts,amssymb}
\usepackage{colortbl}
\definecolor{mygray}{gray}{.9}
\definecolor{mypink}{rgb}{.99,.91,.95}
\definecolor{mycyan}{cmyk}{.3,0,0,0}

\usepackage{caption}
\usepackage[capitalize]{cleveref}
\crefname{section}{Sec.}{Secs.}
\Crefname{section}{Section}{Sections}
\Crefname{table}{Table}{Tables}
\crefname{table}{Tab.}{Tabs.}

\def\cvprPaperID{3393} 
\def\confName{CVPR}
\def\confYear{2022}

\begin{document}

\title{DF-GAN: A Simple and Effective Baseline for Text-to-Image Synthesis}

\author{Ming Tao\textsuperscript{1} \quad 
Hao Tang\textsuperscript{2} \quad
Fei Wu\textsuperscript{1} \quad
Xiaoyuan Jing\textsuperscript{3} \quad
Bing-Kun Bao\textsuperscript{1}\thanks{Corresponding Author} \quad
Changsheng Xu\textsuperscript{4,5,6} \\
\textsuperscript{1}Nanjing University of Posts and Telecommunications \quad  
\textsuperscript{2}CVL, ETH Zürich \quad 
\textsuperscript{3}Wuhan University \\
\textsuperscript{4}Peng Cheng Laboratory \quad
\textsuperscript{5}University of Chinese Academy of Sciences \\
\textsuperscript{6}NLPR, Institute of Automation, CAS \\
{\tt\small bingkunbao@njupt.edu.cn} \\
}

\maketitle
\begin{abstract}

Synthesizing high-quality realistic images from text descriptions is a challenging task. Existing text-to-image Generative Adversarial Networks generally employ a stacked architecture as the backbone yet still remain three flaws. First, the stacked architecture introduces the entanglements between generators of different image scales. Second, existing studies prefer to apply and fix extra networks in adversarial learning for text-image semantic consistency, which limits the supervision capability of these networks. Third, the cross-modal attention-based text-image fusion that widely adopted by previous works is limited on several special image scales because of the computational cost. To these ends, we propose a simpler but more effective Deep Fusion Generative Adversarial Networks (DF-GAN).
To be specific, we propose:
(i) a novel one-stage text-to-image backbone that directly synthesizes high-resolution images without entanglements between different generators,
(ii) a novel Target-Aware Discriminator composed of Matching-Aware Gradient Penalty and One-Way Output, which enhances the text-image semantic consistency without introducing extra networks,
(iii) a novel deep text-image fusion block, which deepens the fusion process to make a full fusion between text and visual features.
Compared with current state-of-the-art methods, our proposed DF-GAN is simpler but more efficient to synthesize realistic and text-matching images and achieves better performance on widely used datasets. 
Code is available at \url{https://github.com/tobran/DF-GAN}.

\end{abstract}

\section{Introduction}
\label{sec:intro}

The last few years have witnessed the great success of Generative Adversarial Networks (GANs) \cite{goodfellow2014generative} for a variety of applications \cite{liu2021generative,wang2020deep,cheng2021fashion}. 
Among them, text-to-image synthesis is one of the most important applications of GANs. It aims to generate realistic and text-consistent images from the given natural language descriptions. Due to its practical value, text-to-image synthesis has become an active research area recently~\cite{yuan2019ckd, hong2018inferring, li2020exploring, cheng2020rifegan, ramesh2021zero, li2019controllable, qiao2019learn, qiao2019mirrorgan, zhu2019dm, yin2019semantics, li2019object, gou2020segattngan}.

\begin{figure}[t] \small
  \centering
  \includegraphics[width=\linewidth]{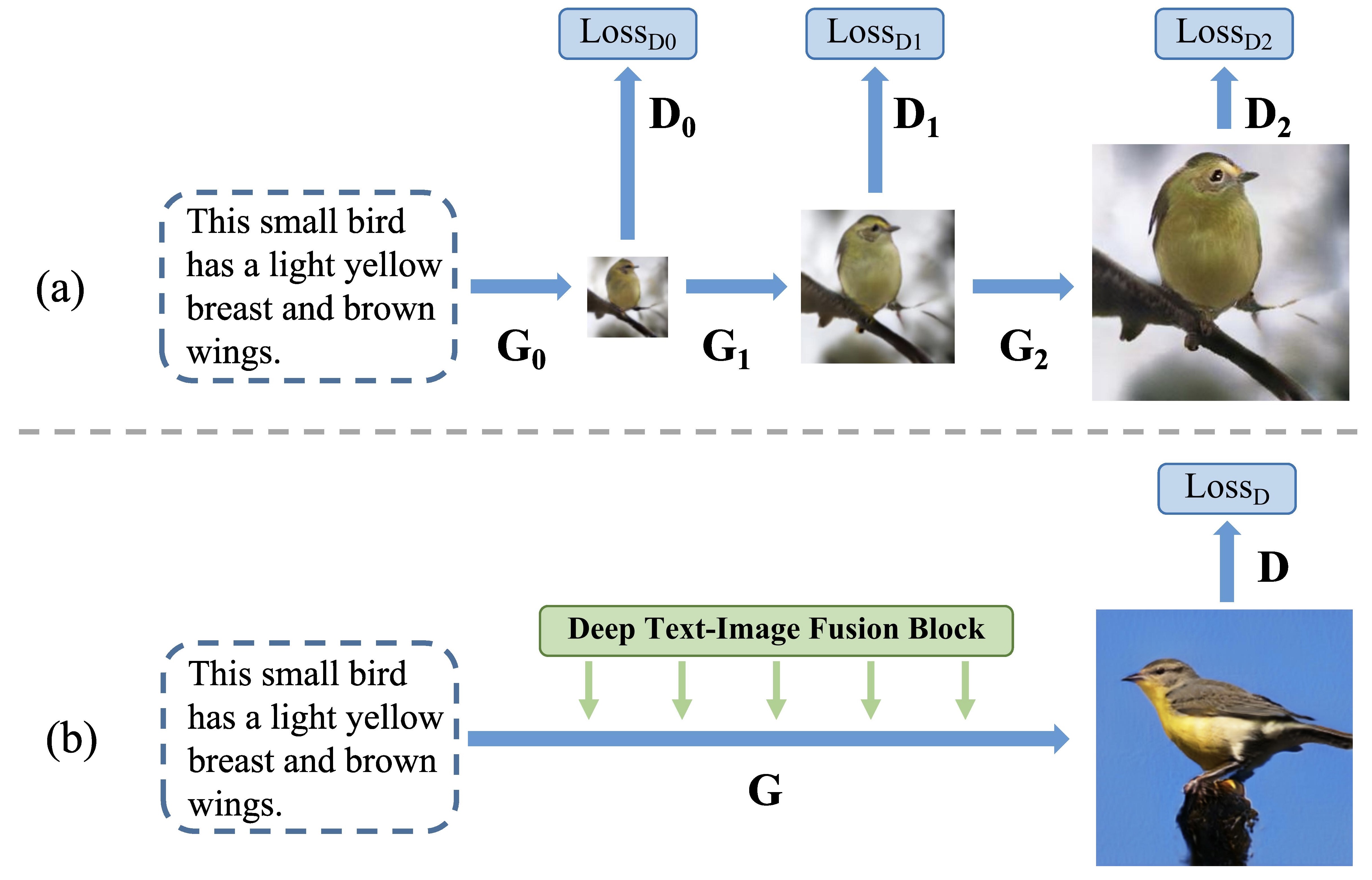}
  \caption{(a) Existing text-to-image models stack multiple generators to generate high-resolution images. (b) Our proposed DF-GAN generates high-quality images directly and fuses the text and image features deeply by our deep text-image fusion blocks.}
  \label{fig1}
  \vspace{-0.4cm}
\end{figure}

Two major challenges for text-to-image synthesis are the authenticity of the generated image, and the semantic consistency between the given text and the generated image.
Due to the instability of the GAN model, most recent models adopt the stacked architecture \cite{zhang2017stackgan, zhang2018stackgan} as the backbone to generate high-resolution images.
They employ cross-modal attention to fuse text and image features~\cite{reed2016generative, zhang2017stackgan, zhang2018stackgan, xu2018attngan, zhu2019dm} and then introduce DAMSM network \cite{xu2018attngan}, cycle consistency \cite{qiao2019mirrorgan}, or Siamese network \cite{yin2019semantics} to ensure the text-image semantic consistency by extra networks.

Although impressive results have been presented by previous works \cite{li2019controllable, qiao2019learn, qiao2019mirrorgan, zhu2019dm, yin2019semantics, li2019object,gou2020segattngan}, there still remain three problems. 
First, the stacked architecture \cite{zhang2017stackgan} introduces entanglements between different generators, and this makes the final refined images look like a simple combination of fuzzy shape and some details.
As shown in Figure~\ref{fig1}(a), the final refined image has a fuzzy shape synthesized by $G_0$, coarse attributes (e.g., eye and beak) synthesized by $G_1$, and fine-grained details (e.g., eye reflection) added by $G_2$.
The final synthesized image looks like a simple combination of visual features from different image scales.
Second, existing studies usually fix the extra networks \cite{xu2018attngan,qiao2019mirrorgan} during the adversarial training, making these networks easily fooled by the generator to synthesize adversarial features \cite{yu2020defense,park2021benchmark}, thereby weakening their supervision power on semantic consistency.
Third, cross-modal attention \cite{xu2018attngan} can not make full use of text information.
They can only be applied two times on $64{\times}64$ and $128{\times}128$ image features due to its high computational cost.
It limits the effectiveness of the text-image fusion process and makes the model hard to extend to higher-resolution image synthesis.

To address the above issues, we propose a novel text-to-image generation method named Deep Fusion Generative Adversarial Network (DF-GAN). 
For the first issue, we replace the stacked backbone with a one-stage backbone.
It is composed of hinge loss~\cite{zhang2019self} and residual networks~\cite{he2016deep} which stabilizes the GAN training process to synthesize high-resolution images directly.
Since there is only one generator in the one-stage backbone, it avoids the entanglements between different generators.

For the second issue, we design a Target-Aware Discriminator composed of Matching-Aware Gradient Penalty (MA-GP) and One-Way Output to enhance the text-image semantic consistency. 
MA-GP is a regularization strategy on the discriminator.
It pursues the gradient of discriminator on target data (real and text-matching image) to be zero.
Thereby, the MA-GP constructs a smooth loss surface at real and matching data points which further promotes the generator to synthesize text-matching images.
Moreover, considering that the previous Two-Way Output slows down the convergence process of the generator under MA-GP, we replace it with a more effective One-Way Output.

For the third issue, we propose a Deep text-image Fusion Block (DFBlock) to fuse the text information into image features more effectively. 
The DFBlock consists of several Affine Transformations \cite{perez2018film}.
The Affine Transformation is a lightweight module that manipulates the visual feature maps through channel-wise scaling and shifting operation.
Stacking multiple DFBlocks at all image scales deepens the text-image fusion process and makes a full fusion between text and visual features.

Overall, our contributions can be summarized as follows:
\begin{itemize}
    \item We propose a novel one-stage text-to-image backbone that can synthesize high-resolution images directly without entanglements between different generators.
    \item We propose a novel Target-Aware Discriminator composed of Matching-Aware Gradient Penalty (MA-GP) and One-Way Output. It significantly enhances the text-image semantic consistency without introducing extra networks.
    \item We propose a novel Deep text-image Fusion Block (DFBlock), which fully fuses text and visual features more effectively and deeply.
    \item Extensive qualitative and quantitative experiments on two challenging datasets demonstrate that the proposed DF-GAN outperforms existing state-of-the-art text-to-image models.
\end{itemize}

\section{Related Work}
Generative Adversarial Networks (GANs) \cite{goodfellow2014generative} are an attractive framework that can be used to mimic complex real-world distributions by solving a min-max optimization problem between a generator and discriminator~\cite{zhang2019self,karras2019style,karras2020analyzing,tang2020xinggan}.
For instance, Reed \emph{et al.} first applied the conditional GAN to generate plausible images from text descriptions \cite{reed2016generative, reed2016learning}. 
StackGAN \cite{zhang2017stackgan, zhang2018stackgan} generates high-resolution images by stacking multiple generators and discriminators and provides the text information to the generator by concatenating text vectors as well as the input noises. 
Next, AttnGAN \cite{xu2018attngan} introduces the cross-modal attention mechanism to help the generator synthesize images with more details. 
MirrorGAN \cite{qiao2019mirrorgan} regenerates text descriptions from generated images for text-image semantic consistency \cite{zhu2017unpaired}. 
SD-GAN \cite{yin2019semantics} employs the Siamese structure \cite{varior2016gated,varior2016siamese} to distill the semantic commons from texts for image generation consistency. 
DM-GAN \cite{zhu2019dm} introduces the Memory Network \cite{gulcehre2018dynamic,weston2014memory} to refine fuzzy image contents when the initial images are not well generated in stacked architecture.
Recently, some large transformer-based text-to-image methods~\cite{ramesh2021zero,lin2021m6,ding2021cogview} show excellent performance on complex image synthesis.
They tokenize the images and take the image tokens and word tokens to make auto-regressive training by a unidirectional Transformer \cite{radford2019language,brown2020language}.

Our DF-GAN is much different from previous methods.
First, it generates high-resolution images directly by a one-stage backbone.
Second, it adopts a Target-Aware Discriminator to enhance text-image semantic consistency without introducing extra networks.
Third, it fuses text and image features more deeply and effectively through a sequence of DFBlocks.
Compared with previous models, our DF-GAN is much simpler but more effective in synthesizing realistic and text-matching images.

\begin{figure*}[t]
  \centering
  \includegraphics[width=0.97\linewidth]{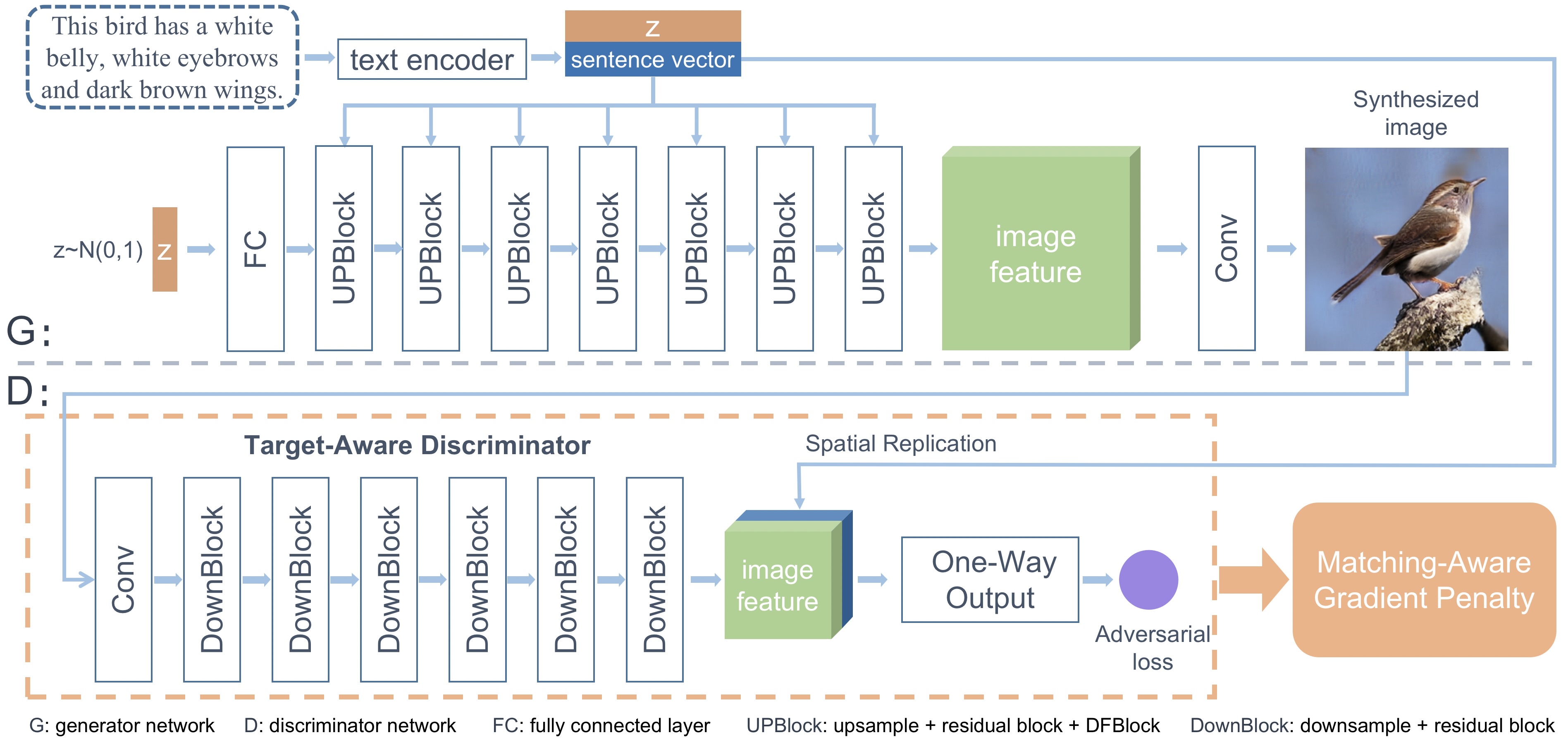}
  \caption{The architecture of the proposed DF-GAN for text-to-image synthesis. DF-GAN generates high-resolution images directly by one pair of generator and discriminator and fuses the text information and visual feature maps through multiple Deep text-image Fusion Blocks (DFBlock) in UPBlocks.  Armed with Matching-Aware Gradient Penalty (MA-GP) and One-Way Output, our model can synthesize more realistic and text-matching images.}
  \label{fig2}
  \vspace{-0.4cm}
\end{figure*}


\section{The Proposed DF-GAN}

In this paper, we propose a simple model for text-to-image synthesis named Deep Fusion GAN (DF-GAN).
To synthesize more realistic and text-matching images, we propose: 
(i) a novel one-stage text-to-image backbone that can synthesize high-resolution images directly without visual feature entanglements.
(ii) a novel Target-Aware Discriminator composed of Matching-Aware Gradient Penalty (MA-GP) and One-Way Output, which enhances the text-image semantic consistency without introducing extra networks.
(iii) a novel Deep text-image Fusion Block (DFBlock), which more fully fuses text and visual features.

\subsection{Model Overview}
The proposed DF-GAN is composed of a generator, a discriminator, and a pre-trained text encoder as shown in Figure \ref{fig2}. 
The generator has two inputs, a sentence vector encoded by text encoder and a noise vector sampled from the Gaussian distribution to ensure the diversity of the generated images. 
The noise vector is first fed into a fully connected layer and reshaped.
We then apply a series of UPBlocks to upsample the image features. 
The UPBlock is composed of an upsample layer, a residual block, and DFBlocks to fuse the text and image features during the image generation process. 
Finally, a convolution layer converts image features into images.

The discriminator converts images into image features through a series of DownBlocks.
Then the sentence vector will be replicated and concatenated with image features. 
An adversarial loss will be predicted to evaluate the visual realism and semantic consistency of inputs. 
By distinguishing generated images from real samples, the discriminator promotes the generator to synthesize images with higher quality and text-image semantic consistency.

The text encoder is a bi-directional Long Short-Term Memory (LSTM) \cite{schuster1997bidirectional} that extracts semantic vectors from the text description.
We directly use the pre-trained model provided by AttnGAN \cite{xu2018attngan}.

\subsection{One-Stage Text-to-Image Backbone}
Since the instability of the GAN model, previous text-to-image GANs usually employ stacked architecture \cite{zhang2017stackgan,zhang2018stackgan} to generate high-resolution images from low-resolution ones. 
However, the stacked architecture introduces entanglements between different generators, and it makes the final refined images look like a simple combination of fuzzy shape and some details (see Figure~\ref{fig1}(a)).

Inspired by recent studies on unconditional image generation \cite{lim2017geometric,zhang2019self}, we propose a one-stage text-to-image backbone that can synthesize high-resolution images directly by a single pair of generator and discriminator. 
We employ the hinge loss \cite{lim2017geometric} to stabilize the adversarial training process.
Since there is only one generator in the one-stage backbone, it avoids the entanglements between different generators.
As the single generator in our one-stage framework needs to synthesize high-resolution images from noise vectors directly, it must contain more layers than previous generators in stacked architecture. 
To train these layers effectively, we introduce residual networks \cite{he2016deep} to stabilize the training of deeper networks.
The formulation of our one-stage method with hinge loss \cite{lim2017geometric} is as follows:
\begin{equation}
 \begin{aligned}
 L_D = &-\mathbb{E}_{x \sim \mathbb{P}_{r}}[min(0,-1+D(x,e))]\\
       &-(1/2)\mathbb{E}_{G(z)\sim \mathbb{P}_{g}}[min(0,-1-D(G(z),e))]\\
       &-(1/2)\mathbb{E}_{x \sim \mathbb{P}_{mis}}[min(0,-1-D(x,e))]\\
 L_G = &-\mathbb{E}_{G(z)\sim \mathbb{P}_{g}}[{D(G(z),e)}]
 \end{aligned} 
\end{equation}
where $z$ is the noise vector sampled from Gaussian distribution; $e$ is the sentence vector; $\mathbb{P}_{g}$, $\mathbb{P}_{r}$, $\mathbb{P}_{mis}$ denote the synthetic data distribution, real data distribution, and mismatching data distribution, respectively.

\subsection{Target-Aware Discriminator}

In this section, we detailed the proposed Target-Aware Discriminator, which is composed of Matching-Aware Gradient Penalty (MA-GP) and One-Way Output.
The Target-Aware Discriminator promotes the generator to synthesize more realistic and text-image semantic-consistent images.

\begin{figure*}[t] \small
  \centering
  \includegraphics[width=\linewidth]{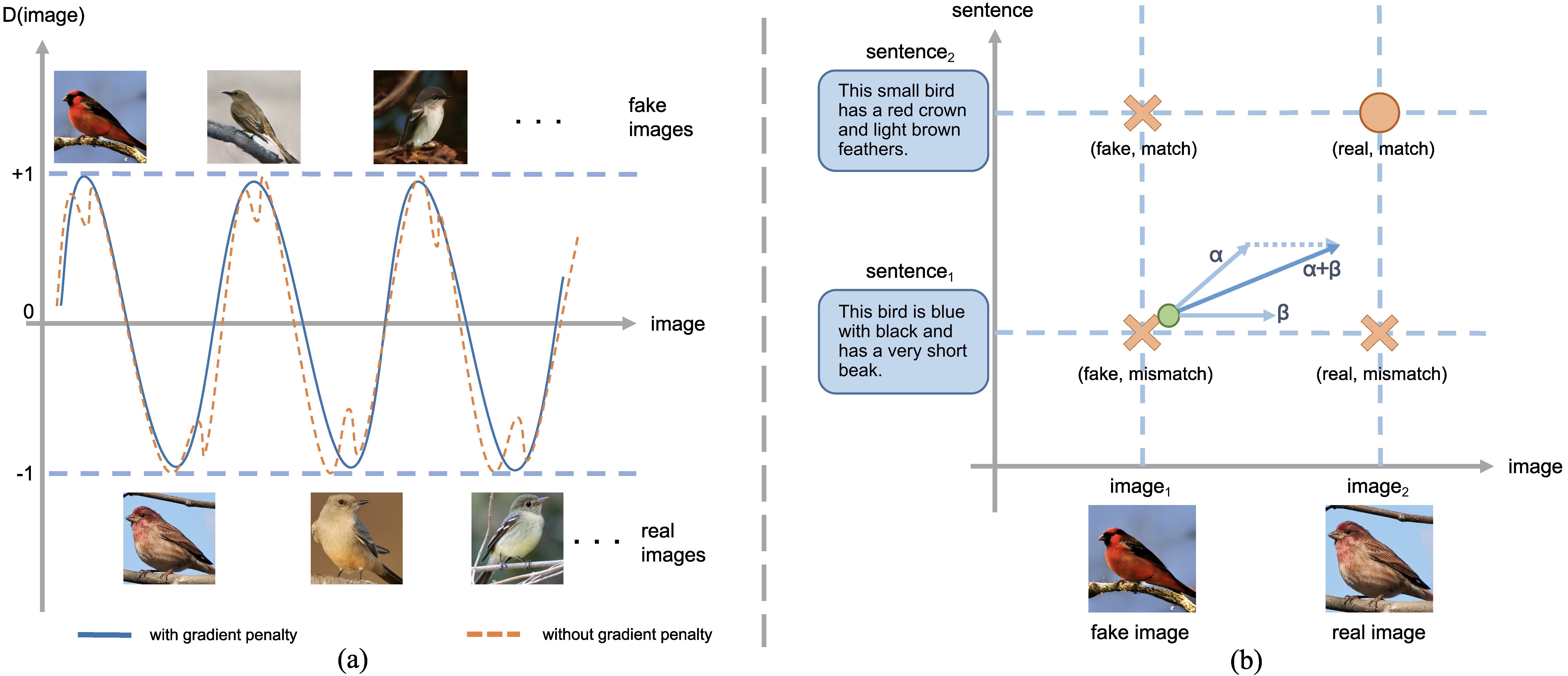}
  \caption{(a) A comparison of loss landscape before and after applying gradient penalty. The gradient penalty smooths the discriminator loss surface which is helpful for generator convergence. (b) A diagram of MA-GP. The data point (real, match) should be applied MA-GP.}
  \label{fig3}
  \vspace{-0.4cm}
\end{figure*}

\subsubsection{Matching-Aware Gradient Penalty}

The Matching-Aware zero-centered Gradient Penalty (MA-GP) is our newly designed strategy to enhance text-image semantic consistency. 
In this subsection, we first show the unconditional gradient penalty \cite{mescheder2018training} from a novel and clear perspective, then extend it to our MA-GP for the text-to-image generation task.

As shown in Figure~\ref{fig3}(a), in unconditional image generation, the target data (real images) correspond to a low discriminator loss.
Correspondingly, the synthetic images correspond to a high discriminator loss. 
The hinge loss limits the range of discriminator loss between -1 and 1. 
The gradient penalty on real data will reduce the gradient of the real data point and its vicinity.
The surface of the loss function around the real data point is then smoothed which is helpful for the synthetic data point to converge to the real data point. 

Based on the above analysis, we find that the gradient penalty on target data constructs a better loss landscape to help the generator converge. 
By leveraging the view into the text-to-image generation.
As shown in Figure~\ref{fig3}(b), in text-to-image generation, the discriminator observes four kinds of inputs: synthetic images with matching text (fake, match), synthetic images with mismatched text (fake, mismatch), real images with matching text (real, match), real images with mismatched text (real, mismatch). 
For text-visual semantic consistency, we tend to apply gradient penalty on the text-matching real data, the target of text-to-image synthesis. 
Therefore, in MA-GP, the gradient penalty should be applied on real images with matching text. 
The whole formulation of our model with MA-GP is as follows:

\begin{equation}
 \begin{aligned}
 L_D = &-\mathbb{E}_{x \sim \mathbb{P}_{r}}[min(0,-1+D(x,e))]\\
       &-(1/2)\mathbb{E}_{G(z)\sim \mathbb{P}_{g}}[min(0,-1-D(G(z),e))]\\
       &-(1/2)\mathbb{E}_{x \sim \mathbb{P}_{mis}}[min(0,-1-D(x,e))]\\
       &+k\mathbb{E}_{x \sim \mathbb{P}_{r}}[(\|\nabla_{x}D(x,e)\|+\|\nabla_{e}D(x,e)\|)^{p}]\\
 L_G = &-\mathbb{E}_{G(z)\sim \mathbb{P}_{g}}[{D(G(z),e)}]
 \end{aligned} 
\end{equation}
where $k$ and $p$ are two hyper-parameters to balance the effectiveness of gradient penalty. 

By using the MA-GP loss as a regularization on the discriminator, our model can better converge to the text-matching real data, therefore synthesizing more text-matching images.
Besides, since the discriminator is jointly trained in our network, it prevents the generator from synthesizing adversarial features of the fixed extra network.
Moreover, since MA-GP does not incorporate any extra networks for text-image consistency and the gradients are already computed by back propagation process, the only computation introduced by our proposed MA-GP is the gradient summation, which is more computational friendly than extra networks.

\subsubsection{One-Way Output}

\begin{figure}[t] \small
  \centering
  \includegraphics[width=\linewidth]{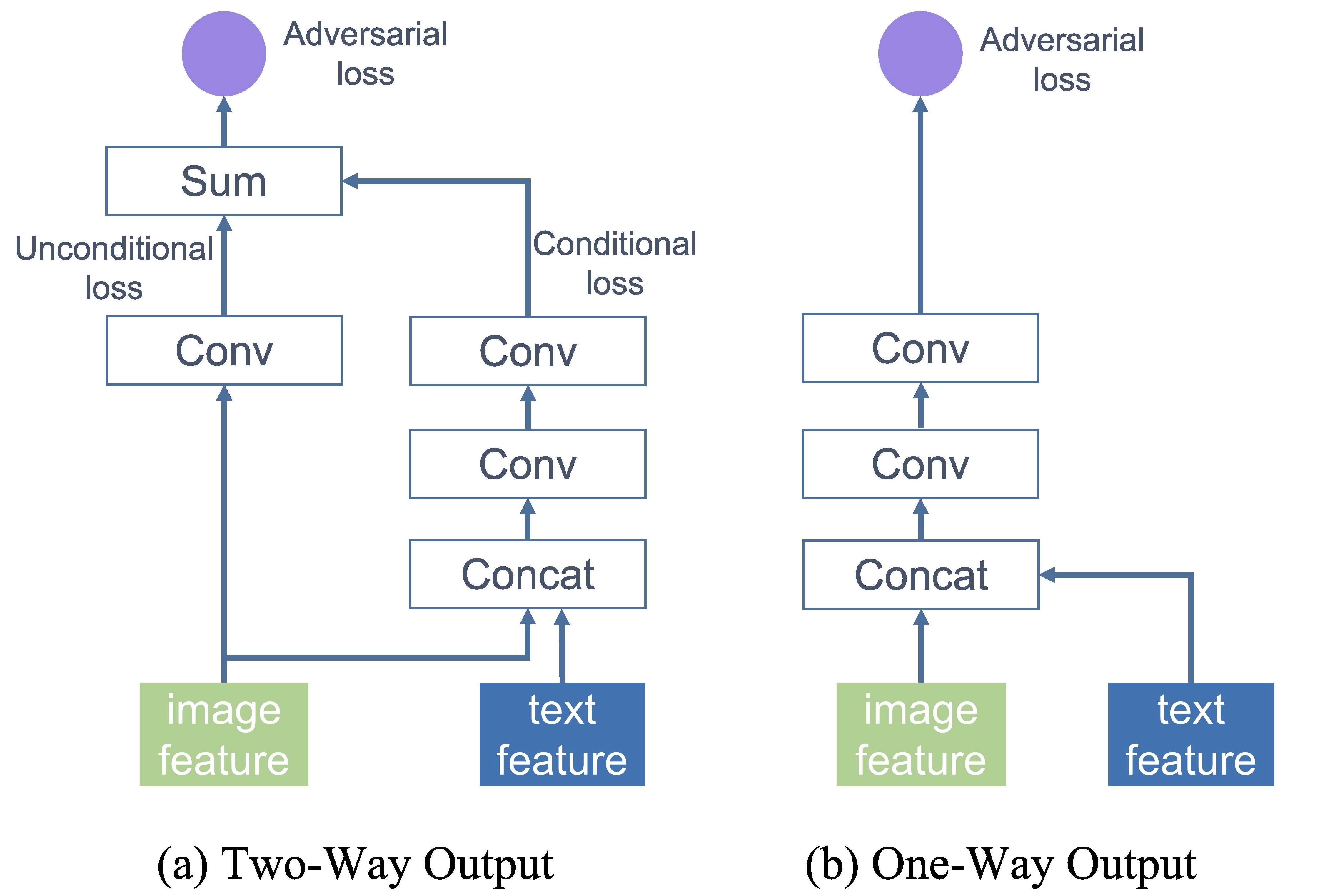}
  \caption{Comparison between Two-Way Output and our One-Way Output. (a) The Two-Way Output predicts conditional loss and unconditional loss and sums them up as the final adversarial loss. (b) Our One-Way Output predicts the whole adversarial loss directly.}
  \label{fig4}
    \vspace{-0.4cm}
\end{figure}

In the previous text-to-image GANs \cite{zhang2017stackgan,zhang2018stackgan,xu2018attngan}, image features extracted by discriminator are usually used in two ways (Figure \ref{fig4}(a)): 
one determines whether the image is real or fake, the other concatenates the image feature and sentence vector to evaluate text-image semantic consistency. 
Correspondingly, the unconditional loss and the conditional loss are computed in these models. 

However, it is shown that the Two-Way Output weakens the effectiveness of MA-GP and slows down the convergence of the generator. 
Concretely, as depicted in Figure \ref{fig3}(b), the conditional loss gives a gradient $\bm{\alpha}$ pointing to the real and matching inputs after back propagation, while the unconditional loss gives a gradient $\bm{\beta}$ only pointing to the real images. 
However, the direction of the final gradient which just simply sums up $\alpha$ and $\beta$ does not point to the real and matching data points as we expected. 
Since the target of the generator is to synthesize real and text-matching images, the final gradient with deviation cannot well achieve text-image semantic consistency and slows down the convergence process of the generator.

Therefore, we propose the One-Way Output for text-to-image synthesis. 
As shown in Figure~\ref{fig4}(b), our discriminator concatenates the image feature and sentence vector, then outputs only one adversarial loss through two convolution layers. 
Through the One-Way Output, we are able to make the single gradient $\bm{\alpha}$ pointed to the target data points (real and match) directly, which optimize and accelerate the convergence of the generator.

By combining the MA-GP and the One-Way Output, our Target-Aware Discriminator can guide the generator to synthesize more real and text-matching images.

\begin{figure*}[t] \small
  \centering
  \includegraphics[width=\linewidth]{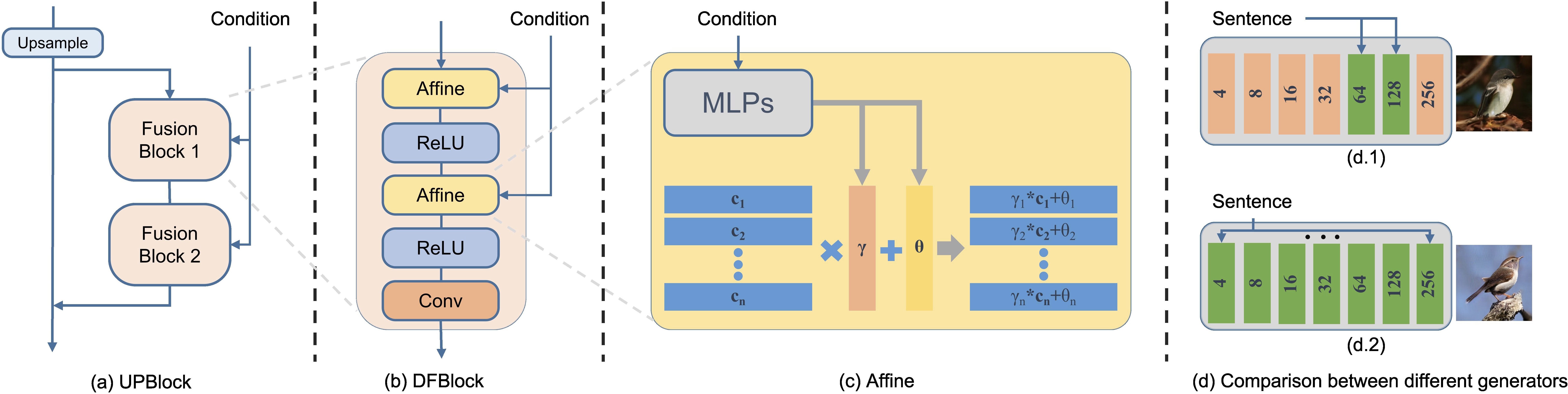}
  \caption{(a) A typical UPBlock in the generator network. The UPBlock upsamples the image features and fuses text and image features by two Fusion Blocks. (b) The DFBlock consists of two Affine layers, two ReLU activation layers, and a Convolution layer. (c) The illustration of the Affine Transformation. (d) Comparison between (d.1) the generator with cross-modal attention \cite{xu2018attngan,zhu2019dm} and (d.2) our generator with DFBlock.}
  \label{fig5}\
  \vspace{-0.4cm}
\end{figure*}

\subsection{Efficient Text-Image Fusion}

To fuse text and image features efficiently, we propose a novel Deep text-image Fusion Block (DFBlock).
Compared with previous text-image fusion modules, our DFBlock deepens the text-image fusion process to make a full text-image fusion.

As shown in Figure 2, the generator of our DF-GAN consists of 7 UPBlocks.
A UPBlock contains two Text-Image Fusion blocks.
To fully utilize the text information in fusion, we propose the Deep text-image Fusion Block (DFBlock) which stacks multiple Affine Transformations and ReLU layers in Fusion Block. 
For Affine transformation, as shown in Figure~\ref{fig5}(c), we adopt two MLPs (Multilayer Perceptron) to predict the language-conditioned channel-wise scaling parameters $\bm{\gamma}$ and shifting parameters $\bm{\theta}$ from sentence vector $\bm{e}$, respectively:
\begin{equation}
\bm{\gamma} = MLP_1(\bm{e}),\qquad\bm{\theta} = MLP_2(\bm{e}).
\end{equation}

For a given input feature map $X {\in} \mathbb{R}^{B\times C\times H\times W}$, 
we first conduct the channel-wise scaling operation on $X$ with the scaling parameter $\bm{\gamma}$, then apply the channel-wise shifting operation with the shifting parameter $\bm{\theta}$. 
Such a process can be expressed as follows:
\begin{equation}
AFF(\bm{x_i}|\bm{e})=\gamma_{i}\cdot{\bm{x_{i}}}+\theta_{i},
\label{eq4}
\end{equation}
where $AFF$ denotes the Affine Transformation; $\bm{x_i}$ is the $i^{th}$ channel of visual feature maps; $\bm{e}$ is the sentence vector; $\gamma_i$ and $\theta_i$ are scaling parameter and shifting parameter for the $i^{th}$ channel of visual feature maps. 

The Affine layer expands the conditional representation space of the generator.
However, the Affine transformation is a linear transformation for each channel.
It limits the effectiveness of text-image fusion process.
Thereby, we add a ReLU layer between two Affine layers which brings the nonlinearity into the fusion process.
It enlarges the conditional representation space compared with only one Affine layer.
A larger representation space is helpful for the generator to map different images to different representations according to text descriptions.

Our DFBlock is partly inspired by Conditional Batch Normalization (CBN)\cite{de2017modulating} and Adaptive Instance Normalization (AdaIN)\cite{huang2017arbitrary,karras2019style} which contain the Affine transformation.
However, both CBN and AdaIN employ normalization layers \cite{ioffe2015batch,ulyanov2016instance} which transform the feature maps into the normal distribution.
It generates an opposite effect to the Affine Transformation which is expected to increase the distance between different samples.
It is then unhelpful for the conditional generation process.
To this end, we remove the normalization process.
Furthermore, our DFBlock deepens the text-image fusion process. 
We stack multiple Affine layers and add a ReLU layer between.
It promotes the diversity of visual features and enlarges the representation spaces to represent different visual features according to different text descriptions.

With the deepening of the fusion process, the DFBlock brings two main benefits for text-to-image generation: 
First, it makes the generator more fully exploit the text information when fusing text and image features.
Second, deepening the fusion process enlargers the representation space of the fusion module, which is beneficial to generate semantic consistent images from different text descriptions.

Furthermore, compared with previous text-to-image GANs \cite{zhang2017stackgan,zhang2018stackgan,xu2018attngan,zhu2019dm},
the proposed DFBlock makes our model no longer consider the limitation from image scales when fusing the text and image features.
This is because existing text-to-image GANs generally employ the cross-modal attention mechanism which suffers a rapid growth of computation cost along with the increase of image size.

\section{Experiments}
In this section, we first introduce the datasets, training details, and evaluation metrics used in our experiments, then evaluate DF-GAN and its variants quantitatively and qualitatively. 

\noindent{\bf Datasets.} We follow previous work~\cite{zhang2017stackgan,zhang2018stackgan,xu2018attngan,zhu2019dm,yin2019semantics,qiao2019mirrorgan} and evaluate the proposed model on two challenging datasets, i.e., CUB bird \cite{wah2011caltech} and COCO \cite{lin2014microsoft}.
The CUB dataset contains 11,788 images belonging to 200 bird species. Each bird image has ten language descriptions.
The COCO dataset contains 80k images for training and 40k images for testing. Each image in this dataset has five language descriptions. 

\noindent{\bf Training Details.} We optimize our network using Adam \cite{kingma2014adam} with $\beta_{1}{=}0.0$ and $\beta_{2}{=}0.9$. The learning rate is set to $0.0001$ for the generator and $0.0004$ for the discriminator according to Two Timescale Update Rule (TTUR) \cite{heusel2017gans}. 

\noindent{\bf Evaluation Details.} Following previous works \cite{xu2018attngan, zhu2019dm}, we choose the Inception Score (IS) \cite{salimans2016improved} and Fr\'echet Inception Distance (FID) \cite{heusel2017gans} to evaluate the performance of our network. 
Specifically, IS computes the Kullback-Leibler (KL) divergence between a conditional distribution and marginal distribution. 
Higher IS means higher quality of the generated images, and each image clearly belongs to a specific class. 
FID \cite{heusel2017gans} computes the Fr\'echet distance between the distribution of the synthetic images and real-world images in the feature space of a pre-trained Inception v3 network. 
Contrary to IS, more realistic images have a lower FID.
To compute both IS and FID, each model generates 30,000 images ($256{\times}256$ resolution) from text descriptions randomly selected from the test dataset. 

As stated in the recent works \cite{li2019object, zhang2021dtgan}, the IS cannot evaluate the image quality well on the COCO dataset, which also exists in our proposed method.
Moreover, we find that some GAN-based models \cite{xu2018attngan,zhu2019dm} achieve significant higher IS than Transformer-based large text-to-image models\cite{ramesh2021zero, ding2021cogview} on the COCO dataset, but the visual quality of synthesized images is obviously lower than Transformer-based models\cite{ramesh2021zero, ding2021cogview}.
Thus, we do not compare IS on the COCO dataset. 
In contrast, FID is more robust and aligns human qualitative evaluation on the COCO dataset. 

Moreover, we evaluate the number of parameters (NoP) to compare the model size with current methods.

\begin{figure*}[t] \small
  \centering
  \includegraphics[width=\linewidth]{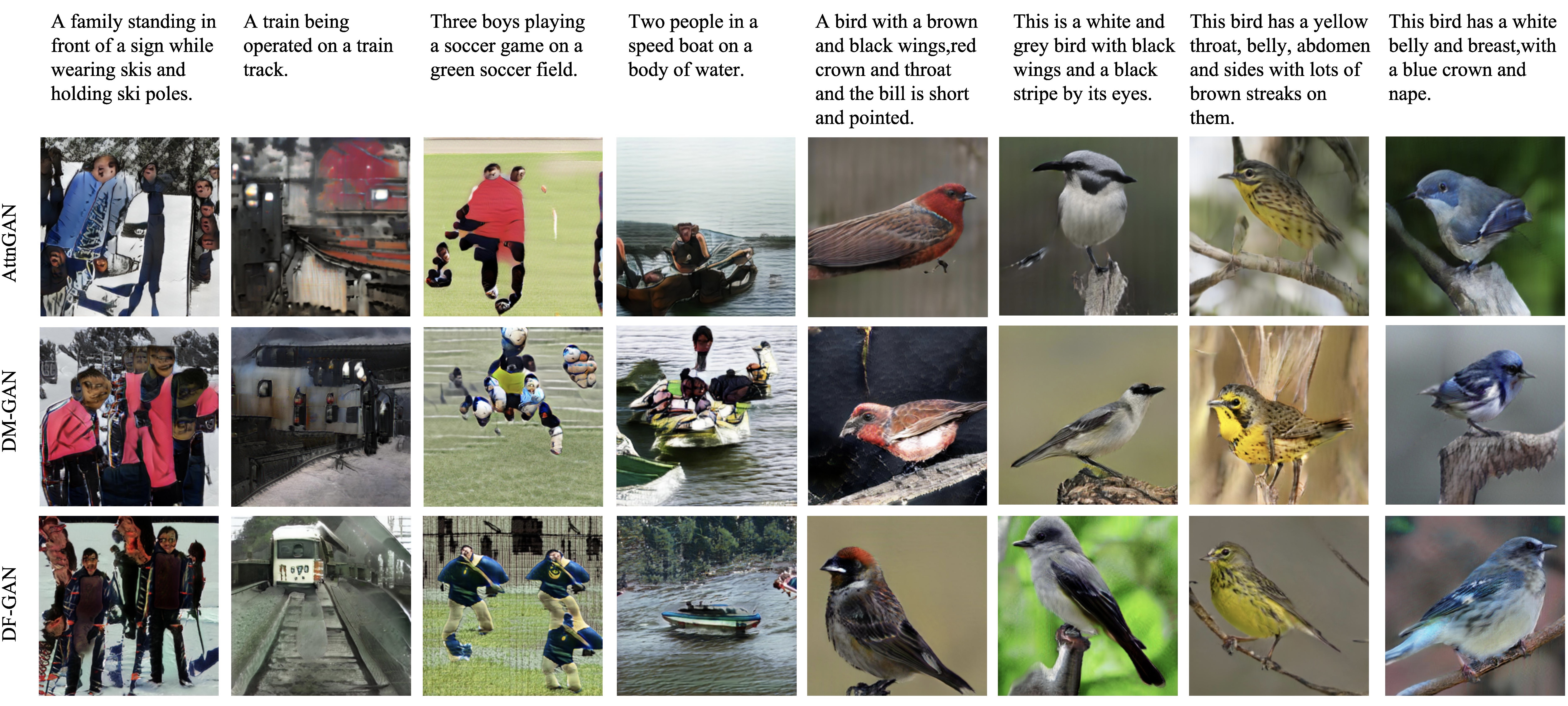}
  \caption{Examples of images synthesized by AttnGAN \cite{xu2018attngan}, DM-GAN \cite{zhu2019dm}, and our proposed DF-GAN conditioned on text descriptions from the test set of COCO and CUB datasets.}
  \label{fig6}
  \vspace{-0.4cm}
\end{figure*}

\begin{table}[t] \small
\centering
\caption{The results of IS, FID and NoP compared with the state-of-the-art methods on the test set of CUB and COCO.}
\begin{tabular}{l|c|c|c|c}
\toprule
\multirow{2}*{Model}                 & \multicolumn{2}{c|}{CUB}                            & \multicolumn{2}{c}{COCO}             \\ 
\cline{2-5}   
                                     & IS $\uparrow$         & FID $\downarrow$            & FID $\downarrow$        & NoP $\downarrow$   \\ \midrule
StackGAN \cite{zhang2017stackgan}    &3.70                   & -                           & -                       & -                  \\
StackGAN++ \cite{zhang2018stackgan}  &3.84                   & -          & -     \\
AttnGAN \cite{xu2018attngan}         & 4.36                  & 23.98                       & 35.49                   & 230M                       \\
MirrorGAN \cite{qiao2019mirrorgan}   &4.56                   & 18.34                       & 34.71                   & -                \\
SD-GAN \cite{yin2019semantics}       &4.67                   & -                           & -                       & -                     \\
DM-GAN  \cite{zhu2019dm}             & 4.75                  & 16.09                       & 32.64                   & 46M                  \\\hline
CPGAN  \cite{liang2020cpgan}          & -                     & -                           & 55.80                    & 318M                    \\
XMC-GAN   \cite{zhang2021cross}     & -                     & -                           & \textcolor{red}{9.30}    & 166M                  \\
DAE-GAN   \cite{ruan2021dae}       & 4.42                   & 15.19                        & 28.12                    & 98M               \\
TIME      \cite{liu2020time}        & \textcolor{cyan}{4.91} & \textcolor{red}{14.30}       & 31.14                    & 120M                  \\ \hline
\rowcolor{mygray}
DF-GAN (Ours)                        & \textcolor{red}{5.10} & \textcolor{cyan}{14.81}     & \textcolor{cyan}{19.32} & \textcolor{red}{19M}   \\ \bottomrule
\end{tabular}
\label{table1}
\vspace{-0.4cm}
\end{table}

\subsection{Quantitative Evaluation}
We compare the proposed method with several state-of-the-art methods, including StackGAN \cite{zhang2017stackgan}, StackGAN++ \cite{zhang2018stackgan}, AttnGAN \cite{xu2018attngan}, MirrorGAN \cite{qiao2019mirrorgan}, SD-GAN \cite{yin2019semantics}, and DM-GAN \cite{zhu2019dm}, which have achieved the remarkable success of text-to-image synthesis by using stacked structures. 
We also compared with more recent models \cite{liang2020cpgan, zhang2021cross, ruan2021dae, liu2020time}.
It should be pointed that recent models always use extra knowledge or supervisions.
For example, CPGAN\cite{liang2020cpgan} uses the extra pretrained YOLO-V3 \cite{redmon2018yolov3}, XMC-GAN \cite{zhang2021cross} uses the extra pretrained VGG-19 \cite{simonyan2014very} and Bert \cite{devlin2018bert}, DAE-GAN \cite{ruan2021dae}  uses extra NLTK POS tagging and manually designs rules for different datasets, and TIME \cite{liu2020time} uses extra 2-D positional encoding.

As shown in Table~\ref{table1}, compared with other leading models, our DF-GAN has a significant smaller Number of Parameters (NoP) but still achieves a competitive performance.
Compared with AttnGAN \cite{xu2018attngan} which employs cross-modal attention to fuse text and image features, our DF-GAN improves the IS metric from 4.36 to 5.10 and decreases the FID metric from 23.98 to 14.81 on the CUB dataset. 
And our DF-GAN decreases FID from 35.49 to 19.32 on the COCO dataset.
Compared with MirrorGAN \cite{qiao2019mirrorgan} and SD-GAN \cite{yin2019semantics} which employ cycle consistency and Siamese network to ensure text-image semantic consistency, our DF-GAN improves IS from 4.56 and 4.67 to 5.10. respectively on the CUB dataset.
Compared with DM-GAN \cite{zhu2019dm} which introduces Memory Network to refine fuzzy image contents, our model also improves IS from 4.75 to 5.10 and decreases FID from 16.09 to 14.81 on CUB, and also decreases FID from 32.64 to 19.32 on the COCO.
Moreover, compared with recent models which introduce extra knowledge, our DF-GAN still achieves a competitive performance.
The quantitative comparisons prove that our model is much simpler but more effective.

\subsection{Qualitative Evaluation}
We also compare the visualization results synthesized by AttnGAN \cite{xu2018attngan}, DM-GAN \cite{zhu2019dm}, and the proposed DF-GAN. 

It can be seen that images synthesized by AttnGAN \cite{xu2018attngan} and DM-GAN \cite{zhu2019dm} in Figure~\ref{fig6} look like a simple combination of fuzzy shape and some visual details (1$^{st}$, 3$^{rd}$, 5$^{th}$, 7$^{th}$, and 8$^{th}$ columns). 
As shown in the 5$^{th}$, 7$^{th}$, and 8$^{th}$ columns, the birds synthesized by AttnGAN \cite{xu2018attngan} and DM-GAN \cite{zhu2019dm} contain wrong shapes. 
Moreover, the images synthesized by our DF-GAN have better object shapes and realistic fine-grained details (e.g., 1$^{st}$, 3$^{rd}$, 7$^{th}$, and 8$^{th}$ columns). 
Besides, the posture of the bird in our DF-GAN result is also more natural (e.g., 7$^{th}$ and 8$^{th}$ columns). 

Comparing the text-image semantic consistency with other models, we find that our DF-GAN can also capture more fine-grained details in text descriptions.
For example, as the results shown in 1$^{st}$, 2$^{th}$, 6$^{th}$ columns in Figure~\ref{fig6}, other models cannot synthesize the ``holding ski poles'', ``train track'', and ``a black stripe by its eyes'' described in the text well, but the proposed DF-GAN can synthesize them more correctly.

\subsection{Ablation Study}

In this section, we conduct ablation studies on the testing set of the CUB dataset to verify the effectiveness of each component in the proposed DF-GAN. 
The components include One-Stage text-to-image Backbone (OS-B), Matching-Aware Gradient Penalty (MA-GP), One-Way Output (OW-O), Deep text-image Fusion Block (DFBlock). 
We also compare our Target-Aware Discriminator with Deep Attentional Multimodal Similarity Model (DAMSM) which is an extra network widely employed in current models \cite{xu2018attngan,yin2019semantics,zhu2019dm}.
We first evaluate the effectiveness of OS-B, MA-GP, and OW-O. 
We conducted a user study to evaluate the text-image semantic consistency (SC), and we asked ten users to score the 100 randomly synthesized images with text descriptions. 
The scores range from 1 (worst) to 5 (best).
The results on the CUB dataset are shown in Table~\ref{table3}.

\begin{table}[t] \small
\centering
\caption{The performance of different components of our model on the test set of CUB.}
\begin{tabular}{l|c|c|c}\toprule
Architecture        &IS $\uparrow$   &FID $\downarrow$      &SC $\uparrow$ \\ \midrule
Baseline            & 3.96           & 51.34                & -\\
OS-B                & 4.11           & 43.45                & 1.46\\
OS-B w/ DAMSM       & 4.28           & 36.72                & 1.79\\
OS-B w/ MA-GP       & 4.46           & 32.52                & 3.55\\
OS-B w/ MA-GP w/ OW-O & \textbf{4.57}  & \textbf{23.16} & \textbf{4.61}\\ \bottomrule
\end{tabular}
\label{table3}
\vspace{-0.4cm}
\end{table}

\noindent{\bf Baseline.} 
Our baseline employs stacked framework and Two-Way Output with the same Adversarial loss as StackGAN\cite{zhang2017stackgan}.
In baseline, the sentence vector is naively concatenated to the input noise and intermediate feature maps.

\noindent{\bf Effect of One-Stage Backbone.} Our proposed OS-B improves IS from 3.96 to 4.11 and decreases FID from 43.45 to 32.52. 
The results demonstrate that our one-stage backbone is more effective than stacked architecture.

\noindent{\bf Effect of MA-GP.} Armed with MA-GP, the model further improves IS to 4.46, SC to 3.55, and decreases FID to 32.52 significantly. 
It demonstrates that the proposed MA-GP can promote the generator to synthesize more realistic and text-image semantic consistent images.

\noindent{\bf Effect of One-Way Output.} The proposed OW-O also improves IS from 4.46 to 4.57, SC from 3.55 to 4.61, and decreases FID from 32.52 to 23.16. 
It also demonstrates that the One-Way Output is more effective than a Two-Way Output in the text-to-image generation task. 

\noindent{\bf Effect of Target-Aware Discriminator.} Compared with DAMSM, our proposed Target-Aware Discriminator composed of MA-GP and OW-O improves IS from 4.28 to 4.57, SC from 1.79 to 4.61, and decreases FID from 36.72 to 23.16. 
The results demonstrate that our Target-Aware Discriminator is superior to extra networks.

\noindent{\bf Effect of DFBlock.} 
We compare our DFBlock with CBN \cite{de2017modulating, miyato2018cgans, brock2018large}, AdaIN\cite{karras2019style} and AFFBlock.
The AFFBlock employs one Affine Transformation layer to fuse text and image features. 
MA-GP GAN is the model that employs One-Stage text-to-image Backbone, Matching-Aware Gradient Penalty, and One-Way Output.
From the results in Table~\ref{table4}, we find that, compared with other fusion methods, concatenation cannot efficiently fuse text and image features.
The comparison among CBN, AdaIN, and AFFBlock proves that Normalization is not essential in Fusion Block, and removing normalization even slightly improves the results. 
The comparison between DFBlock and AFFBlock demonstrates the effectiveness of deepening the text-image fusion process. 
In sum, the comparison results prove the effectiveness of our proposed DFBlock.

\subsection{Limitations}
Although DF-GAN shows superiority in text-to-image synthesis, some limitations must be taken into consideration in future studies. 
First, our model only introduces the sentence-level text information, which limits the ability of fine-grained visual feature synthesis.
Second, introducing pre-trained large language models\cite{devlin2018bert,radford2019language} to provide additional knowledge may further improve the performance.
We will try to address these limitations in our future work.

\begin{table}[t] \small
\centering
\caption{The performance of MA-GP GAN with different modules on the test set of CUB.}
\begin{tabular}{l|c|c}\toprule
Architecture                  &IS$\uparrow$    &FID  $\downarrow$\\\midrule
MA-GP GAN w/ Concat           & 4.57           & 23.16\\ 
MA-GP GAN w/ CBN           & 4.81           & 18.56\\
MA-GP GAN w/ AdaIN            & 4.85           & 17.52\\
MA-GP GAN w/ AFFBLK           & 4.87           & 17.43\\
MA-GP GAN w/ DFBLK (DF-GAN)   & \textbf{5.10}  & \textbf{14.81} \\ \bottomrule
\end{tabular}
\label{table4}
\vspace{-0.4cm}
\end{table}

\section{Conclusion and Future Work}

In this paper, we propose a novel DF-GAN for the text-to-image generation tasks.
We present a one-stage text-to-image Backbone that can synthesize high-resolution images directly without entanglements between different generators.
We also propose a novel Target-Aware Discriminator composed of Matching-Aware Gradient Penalty (MA-GP) and One-Way Output.
It can further enhance the text-image semantic consistency without introducing extra networks.
Besides, we introduce a novel Deep text-image Fusion Block (DFBlock) which fully fuses text and image features more effectively and deeply.
Extensive experiment results demonstrate that our proposed DF-GAN significantly outperforms current state-of-the-art models on the CUB dataset and more challenging COCO dataset. 

\section*{Acknowledgment}

This work was supported by National Key Research and Development Project (No.2020AAA0106200),
the National Nature Science Foundation of China under Grants (No.61936005, 61872424, 62076139, 62176069 and 61933013),
the Natural Science Foundation of Jiangsu Province (Grants No.BK20200037 and BK20210595),
and the Open Research Project of Zhejiang Lab (No.2021KF0AB05).

\clearpage
{\small
\bibliographystyle{ieee_fullname}
\bibliography{main}
}

\end{document}